\pdfoutput=1

\documentclass[11pt]{article}

\usepackage[final]{acl}

\usepackage{times}
\usepackage{latexsym}

\usepackage[T1]{fontenc}

\usepackage[utf8]{inputenc}

\usepackage{microtype}

\usepackage{inconsolata}
\usepackage[utf8]{inputenc}
\usepackage[arabic,farsi,english]{babel}
\usepackage{graphicx}

\title{Matina: A Large-Scale 73B Token Persian Text Corpus}

\author{
  Sara Bourbour Hosseinbeigi \\
  Tarbiat Modares University \\
  \texttt{s.bourbour@modares.ac.ir} \\
  \And
  Fatemeh Taherinezhad \\
  University of Tehran \\
  \texttt{ftaherinezhad@ut.ac.ir} \\
  \And
  Heshaam Faili \\
  University of Tehran \\
  \texttt{hfaili@ut.ac.ir} \\
  \AND
  Hamed Baghbani \\
  University of Tehran \\
  \texttt{baghbani.hamed@ut.ac.ir} \\
  \And
  Fatemeh Nadi \\
  University of Tehran \\
  \texttt{fatemehnadi@ut.ac.ir} \\
  \And
  Mostafa Amiri \\
  University of Tehran \\
  \texttt{mostafa.amiri@ut.ac.ir}
}

\begin{document}
\maketitle
\begin{abstract}
Text corpora are essential for training models used in tasks like summarization, translation, and large language models (LLMs). While various efforts have been made to collect monolingual and multilingual datasets in many languages, Persian has often been underrepresented due to limited resources for data collection and preprocessing. Existing Persian datasets are typically small and lack content diversity, consisting mainly of weblogs and news articles. This shortage of high-quality, varied data has slowed the development of NLP models and open-source LLMs for Persian. Since model performance depends heavily on the quality of training data, we address this gap by introducing the Matina corpus, a new Persian dataset of 72.9B tokens, carefully preprocessed and deduplicated to ensure high data quality. We further assess its effectiveness by training and evaluating transformer-based models on key NLP tasks. Both the dataset and preprocessing codes are publicly available\footnote{\href{https://github.com/FTaheriN/Matina-Text-Preprocessing}{https://github.com/FTaheriN/Matina-Text-Preprocessing}}, enabling researchers to build on and improve this resource for future Persian NLP advancements.
    \end{abstract}
    
\begin{table*}
  \centering
  \begin{tabular}{lll}
    \hline
    \textbf{Component}           & \textbf{Number of tokens} & \textbf{Mean document length} \\
    \hline
    Books           &    2,842,128,225   &   162,648.9   \\
    Papers          &    3,547,046,981   &   10,620.5    \\
    Social Media    &    2,143,415,349   &   351.6       \\
    Web Crawled     &    14,782,414,716     &   749.7        \\
    CulturaX FA     &    20,469,778,795     &   1,124.8       \\
    MADLAD-400 FA   &    29,131,569,264     &   1,352.96      \\
    \hline
    \textbf{Matina} &    72,916,353,330  &   1,106.5     \\
    \hline
  \end{tabular}
  \caption{\label{table:corpus-size}
    Overview of components in Matina Corpus. Tokens are counted by the Llama 3.1 \citep{dubey2024llama3.1} tokenizer. 
  }
\end{table*}

\section{Introduction}
Since the introduction of the transformer architecture \citep{vaswani2017attention}, natural language processing (NLP) has advanced rapidly, transforming many language-related tasks. Transformer-based models, like BERT \citep{devlin2018bert} and GPT-2 \citep{radford2018improving}, initially focused on tasks like sentiment analysis, translation, and summarization. However, with the development of large-scale language models (LLMs), such as GPT-3 \citep{brown2020language} and later models \citep{touvron2023llama, le2023bloom, bai2023qwen, yang2024qwen2}, the research shifted towards more complex tasks, including generalization, creative problem-solving, and critical thinking.

The performance of these models, in both basic and advanced tasks, isn't just about model size or computational power—it's also heavily influenced by the quality and amount of training data. As a result, a lot of effort has gone into large-scale data collection and preprocessing \citep{gao2020pile, laurenccon2022roots, penedo2023refinedweb} to improve model capabilities and generalization.

While English dominates NLP research, there has been a growing effort to curate multilingual datasets \citep{wenzek2019ccnet, laurenccon2022roots, nguyen2023culturax, kudugunta2024madlad} and develop models capable of understanding multiple languages \citep{le2023bloom, touvron2023llama, yang2024qwen2}.

Despite Persian being widely spoken, it remains underrepresented in NLP research. Although both conventional models and LLMs can process Persian, their performance is often suboptimal, mainly because of the limited availability and poor quality of existing data. Persian text data is predominantly sourced from news websites and blogs, which often lack formal or factual content. Moreover, no standardized preprocessing pipeline exists to ensure the high quality of Persian datasets at the same level as those available for other languages.

To address this gap, we introduce the Matina Corpus, a 72.9 billion token Persian dataset designed for training language models. Unlike other Persian datasets \citep{targoman, sabeti2018mirastext}, the Matina Corpus has undergone a rigorous and well-designed preprocessing pipeline and a comprehensive deduplication process to ensure its high quality. The dataset includes not only publicly available Persian datasets but also introduces newly collected sources to ensure greater diversity and the inclusion of factual information. The diverse sources in the dataset make it suitable both for training large language models and for a variety of downstream tasks that require clean, high-quality Persian data.

The Matina Corpus includes Persian sections from Madlad \citep{kudugunta2024madlad}, CulturaX \citep{nguyen2023culturax}, and the most recent Persian Wikipedia update.  Each data source was processed differently, based on heuristics derived from careful evaluation and observation of the content. To ensure quality and avoid redundancy, deduplication was applied to related chunks of documents rather than across the entire dataset at once. The final corpus comprises a total of 72.9 B tokens, with an average document length of 1,106.5 across different sources (as summarized in Table~\ref{table:corpus-size}), illustrating both the breadth and depth of the dataset.


The Matina Corpus is designed to enhance Persian NLP by supporting both the pretraining of large language models (LLMs) and the development of smaller models based on transformers and other architectures. It enables various NLP tasks, including text classification, machine translation, and sentiment analysis. To evaluate its impact, we continued the pretraining of XML-RoBERTa \citep{conneau2019unsupervised} on Matina and assessed its performance on sentiment analysis, text emotion detection, and named entity recognition, observing notable improvements over models trained on existing Persian datasets.

Furthermore, integrating this high-quality dataset into multilingual models enhances their Persian language comprehension, helping bridge the resource gap. To measure this effect, we used portions of the corpus to continue pretraining LLaMA 3.1 8B, achieving significant gains in Persian text understanding.

The rest of this paper is structured as follows: We begin by providing an overview of existing large corpora, along with the preprocessing pipelines applied to them, covering English, multilingual, and Persian datasets. Afterward, we introduce our corpus, dividing it into three distinct sections based on content, and offer details on the preprocessing steps we applied. We then assess the dataset’s effectiveness through model training and evaluation. Finally, we analyze the dataset, discuss its limitations, and conclude with a summary of our dataset.

\section{Related Work}
The scope of our dataset encompasses two key dimensions: (1) the preprocessing steps involved in creating large-scale corpora and (2) the development of extensive text corpora in Persian. Accordingly, we divide this section into two parts. First, we review notable large-scale corpora available in languages other than Persian, along with the preprocessing techniques applied to these datasets. Then, we examine and analyze the current state of publicly available Persian corpora.

\subsection{Large-Scale Public Corpora}
Since the early stages of NLP development, there have been efforts to compile large-scale datasets for training models in various downstream tasks, such as sentiment analysis, summarization, and text classification, among others \citep{glockner2018snli, narayan2018xsum, wang2019glue}. With the advent of deep learning models, these efforts have escalated in scope, culminating in the large-scale data collection necessary for training large language models (LLMs). One of the earliest and most significant contributions to the development of large text corpora is Common Crawl \citep{commoncrawl}.

Common Crawl \citep{commoncrawl} is a vast multilingual web corpus that continuously archives webpage data from the Internet. However, Common Crawl contains substantial amounts of extraneous content, including advertisements, navigation bars, and inappropriate materials such as pornography, violence, spam, and sensitive personal information. In response to these issues, datasets like OSCAR \citep{oscar}, C4 \citep{raffel2020c4}, mC4 \citep{xue2020mt5}, The Pile \citep{gao2020pile} RefinedWeb \citep{penedo2023refinedweb}, and FineWeb \citep{penedo2024fineweb}  have been created to provide cleaner and more refined versions of the Common Crawl data.

\citet{oscar} took a parallel method to fastText \citep{athiwaratkun2018fasttext} when preprocessing Common Crawl for better data quality. A linear classifier was used to categorize the WET files for language, followed by a filter for erroneous UTF-8 characters and a hashing approach to remove duplicates. This approach produced a 6.3TB dataset covering 160 languages. Similar pipelines were used to build datasets such as CC-100 \citep{conneau2019cc100} and RedPajama \citep{together2023redpajama}.

Likewise, C4 \citep{raffel2020c4} was constructed from Common Crawl data to train the T5 model. The \href{https://pypi.org/project/langdetect/}{langdetect}\footnote{\href{https://pypi.org/project/langdetect/}{https://pypi.org/project/langdetect/}} tool was employed to filter only English pages. Pages containing inappropriate content, specific keywords, curly brackets (identified as code), or a limited number of lines were removed. Subsequently, a set of heuristics was applied at the line level, including checks for terminal punctuation, JavaScript keywords, and boilerplate text. The documents were then deduplicated using a three-sentence span. Building upon C4 \citep{raffel2020c4}, mC4 \citep{xue2020mt5} expanded the dataset to 107 languages. \href{https://github.com/google/cld3}{Cld}\footnote{\href{https://github.com/google/cld3}{https://github.com/google/cld3}} was used for language classification, and documents with language confidence below 70\% were discarded. As in C4 \citep{raffel2020c4}, deduplication was performed at the final stage.

Due to the limited factual and academic content in previous datasets, The Pile \citep{gao2020pile} introduced 21 additional sources, including books, academic papers, code, and subtitles, alongside Common Crawl data (Pile CC). Each data source was processed using specific heuristics tailored to its structure, and the sources were unified into an English-only dataset of 825 GiB. Similarly, MassiveText \citep{muennighoff2022massiveText} was created to train the Gopher model, drawing from six sources: massiveWeb, books, C4, news, GitHub, and Wikipedia. The web data was filtered based on various criteria, including non-English content, fewer than two English stopwords, excessive bullet points, unsuitable word count or length, and pages with repeated words or phrases. Deduplication was performed using MinHash \citep{broder1997minhash} with Jaccard similarity, producing a multilingual dataset containing 2.53 billion documents.

ROOTS \citep{laurenccon2022roots} is a multilingual corpus that includes 46 natural and 13 programming languages. Although the 1.6TB collection comprises primarily of web-based information, many websites were created through crowdsourcing. Pages were filtered using heuristics and thresholds, with low-quality documents deleted using a pretrained tokenizer. Personal information such as email addresses, phone numbers, and IP addresses were eliminated with regular expressions. To assure data quality, the crowd workers selected language-specific preprocessing methods.

RefinedWeb \citep{penedo2023refinedweb} used a similar preprocessing pipeline to MassiveText, with additional heuristics for document filtering. Starting with web-based data from multiple Common Crawl dumps, English documents were first identified using fastText \citep{athiwaratkun2018fasttext} and then filtered at both the document and line levels. More strict filtering was applied to remove sensitive and adult content. Deduplication was performed using both fuzzy methods and exact substring matching. RedPajamas v2 \citep{together2023redpajama} was created using 84 Common Crawl dumps and the CC-Net \citep{wenzek2019ccnet} preprocessing pipeline, with fuzzy and exact-matching deduplication. This dataset spans five languages and contains 100 billion documents. Building on their earlier dataset, Huggingface introduced FineWeb \citep{penedo2024fineweb} based on 95 Common Crawl snapshots. After following a similar preprocessing procedure to RefinedWeb \citep{penedo2023refinedweb}, additional heuristics and a different deduplication method, derived from extensive ablation studies, were applied. The final processed dataset is 96.4TB in size.

\subsection{Persian Text Corpora}
The rapid development of natural language processing (NLP) has necessitated the creation of diverse, large-scale text corpora across various languages. For Persian, also known as Farsi, the availability of robust datasets is crucial for enhancing language modeling capabilities. However, a significant gap persists in terms of corpora that are sufficiently diverse and preprocessed for effective use in training LLMs. Many existing Persian datasets predominantly feature news content, which does not adequately cover the full spectrum of language use. Despite these limitations, Persian remains a language with rich literary and cultural resources, suggesting a substantial potential for corpus development. 

Several Persian corpora, including the \href{https://github.com/Text-Mining/Persian-Wikipedia-Corpus}{Persian Wikipedia Corpus}\footnote{\href{https://github.com/Text-Mining/Persian-Wikipedia-Corpus}{https://github.com/Text-Mining/Persian-Wikipedia-Corpus}}, \href{https://github.com/miras-tech/MirasText}{MirasText}\footnote{\href{https://github.com/miras-tech/MirasText}{https://github.com/miras-tech/MirasText}}, hmBlogs \citep{khansari2021hmblogsbiggeneralpersian}, Naab \citep{naab}, Targoman \citep{targoman}, have significantly enriched the pool of publicly available Persian data. The \href{https://github.com/Text-Mining/Persian-Wikipedia-Corpus}{Persian Wikipedia Corpus}, with over one million articles, serves as a foundational resource, though its content is mainly formal and factual. \href{https://github.com/miras-tech/MirasText}{MirasText}, covering 2.8 million articles from more than 250 news websites, and Naab \citep{naab}, containing around 15 billion tokens, both contribute vast data but are largely news-centric, which limits content diversity. In contrast, Targoman \citep{targoman} expands the scope by incorporating 65 million documents across weblogs, forums, literature, and educational content, although issues with licensing and accessibility hinder its public use. Additionally, hmBlogs \citep{khansari2021hmblogsbiggeneralpersian} offers a valuable glimpse into colloquial language with 20 million blog posts spanning 15 years, though it requires extensive preprocessing to ensure its consistency and applicability. Additionally, \href{https://github.com/ganjoor}{Ganjoor}\footnote{\href{https://github.com/ganjoor}{https://github.com/ganjoor}} introduces classical Persian poetry from 12 poets, enhancing the stylistic and lexical range of the corpus and providing unique linguistic depth.

Parallel corpora, including TEP: Tehran English-Persian parallel corpus \citep{tiedemann-2012-parallel}, MIZAN \citep{kashefi2020mizanlargepersianenglishparallel}, and the \href{https://github.com/christos-c/bible-corpus}{Bible Corpus}\footnote{\href{https://github.com/christos-c/bible-corpus}{https://github.com/christos-c/bible-corpus}}, further extend the utility of Persian datasets by enabling translation tasks and bilingual language modeling. MIZAN \citep{kashefi2020mizanlargepersianenglishparallel}, containing one million sentence pairings between Persian and English, allows cross-linguistic studies and machine translation. However, the breadth of such corpora is frequently limited.

Standardized preprocessing techniques are required to improve Persian language modeling by filtering non-Farsi words, unifying Arabic and Farsi characters, and removing unnecessary content. These steps are crucial for creating a high-quality, clean corpus, as data quality directly impacts model performance in large language models (LLMs). While some datasets, such as Naab \citep{naab} and hmBlogs \citep{khansari2021hmblogsbiggeneralpersian}, offer preprocessed versions, this is still the exception rather than the norm in Persian corpus development.


\section{Matina Corpus}

\begin{figure}[t]
  \includegraphics[trim=2.5cm 4cm 2.2cm 1.8cm , clip,width=\columnwidth]{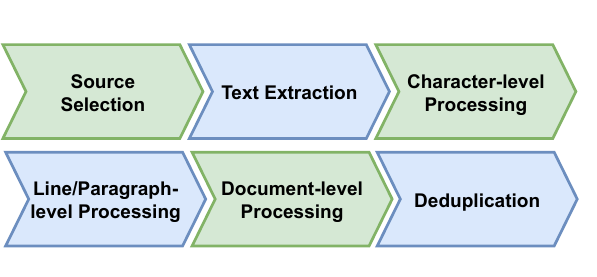}
  \caption{The overall stages of processing pipeline of Matina Corpus.}
  \label{fig:pipeline}

\end{figure}

The Matina corpus is built from a variety of data sources, each of which is processed based on its specific content characteristics. Although these sources are grouped into three main categories, the overall preprocessing pipeline remains consistent, as depicted in Figure~\ref{fig:pipeline}, with variations primarily in hyperparameters. Certain sources, however, demand additional cleaning steps, which are detailed in their respective sections.

\begin{figure}[t]
  \includegraphics[width=\columnwidth]{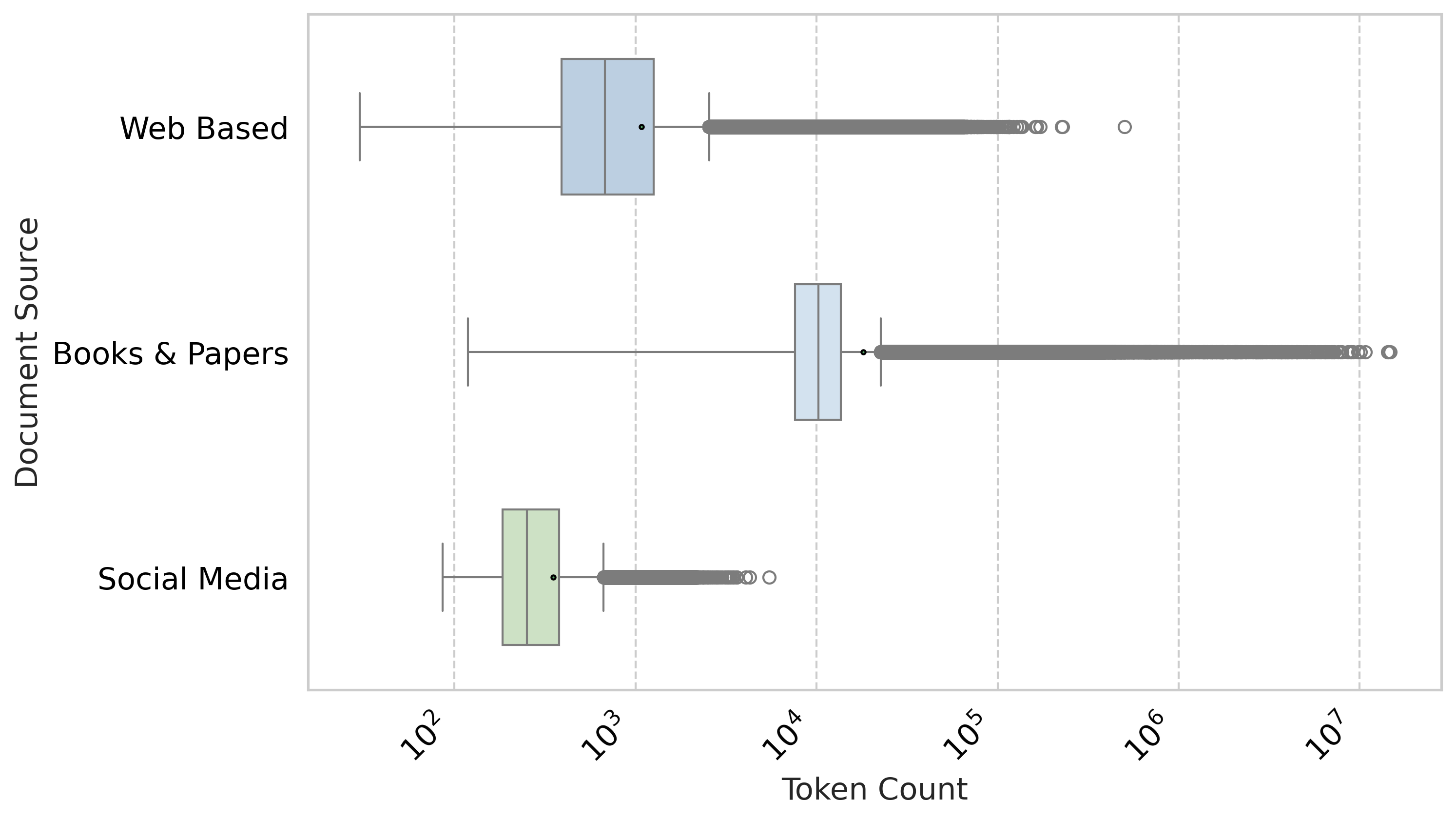}
  \caption{Distribution of document length by source in the Matina Corpus. Length is determined by the log of the number of tokens using Llama3.1 \cite{dubey2024llama3.1} tokenizer.}
  \label{fig:boxplot1}
\end{figure}

Figure~\ref{fig:boxplot1} visualizes the distribution of token counts across documents from each source, using a box plot to illustrate the variance in document length. These three categories—web-based crawled data, crawled books and papers, and social media—form the core of our dataset, each with distinct preprocessing requirements. In this section, we describe the data collection process, the preprocessing techniques applied, and the rationale behind the decisions made throughout these steps.

\subsection{Web-based Crawled Data}
Web crawling is a common and efficient method for collecting data in any language. Websites offer a vast range of valuable information and, given their structured nature and wide availability, can largely be crawled automatically. As a result, web data is frequently used as the primary source for constructing large-scale text datasets. However, while the bulk collection of web data is straightforward, extracting meaningful content from irrelevant elements such as metadata, advertisements, and embedded links remains challenging. Web pages often contain spam-like elements, which complicates the cleaning process and increases the likelihood of errors.

Most web-based datasets begin with basic steps such as text extraction and language detection, often followed by optional URL filtering to exclude content deemed inappropriate or irrelevant. Further preprocessing steps are applied, followed by deduplication to ensure data quality and minimize redundancy. We adopt a similar approach in preprocessing the web data collected for the Matina corpus.

Matina's web-based data is divided into two parts: data crawled by our team and data taken from two public databases using the Common Crawl \citep{commoncrawl} dataset. This dual-source strategy uses both proprietary and publically available data to increase the corpus's breadth and diversity.

In any language, certain domains are recognized for their reliability and high-quality information. We identified such domains in Persian and crawled them to extract relevant textual content. This step helped minimize the inclusion of irrelevant elements such as advertisements, tags, or comments. Text extracted from headings and paragraphs was merged to form unified documents, with additional informative fields (e.g., summaries or subheadings) incorporated as metadata, if available. Because these domains were manually selected, language detection and URL filtering were unnecessary. We also ensured that the selected URLs did not contain harmful, sensitive, or adult content.

For the public datasets, Madlad-400 \citep{kudugunta2024madlad} and CulturaX \citep{nguyen2023culturax}, the initial preprocessing steps—such as language detection, text extraction, and URL filtering—had already been completed by the dataset providers. These datasets also included filters for toxic or harmful content, which allowed us to directly proceed to the next stages of preprocessing. While both datasets applied generic filters—such as language mismatch detection, character ratio checks, and word/sentence length thresholds, these filters were not language-specific. Therefore, we processed data from these sources similarly to the web data we crawled ourselves. After applying the processing on data sourced from web and the public datasets, there remained 64.3B tokens with an average document length of 1,141.8 tokens. 

After inspecting samples from various domains, we defined heuristic functions to modify documents and remove those deemed irrelevant. These heuristics were inspired by preprocessing pipelines adopted in BLOOM \citep{le2023bloom}, MassiveText \citep{muennighoff2022massiveText}, and RefineWeb \citep{penedo2023refinedweb}, but we tailored them to the specific characteristics of our data and added multiple other processing functions. 

Our preprocessing pipeline for web-based data encompasses three primary stages: character-level processing, line and paragraph-level processing, and document-level processing. Each stage employs a series of targeted operations to enhance data quality, ensure linguistic consistency, and eliminate redundancies.  Appendix~\ref{sec:appendixA} provides a full explanation of each step in the preprocessing and deduplication procedures.  

\textbf{Character-level processing} involves normalizing Persian characters, mapping symbols and numbers to their Persian equivalents, limiting the occurrence of repeated characters, standardizing newline characters, and removing non-standard Unicode symbols. This stage ensures that the text adheres to consistent encoding standards and minimizes the presence of corrupted or irrelevant characters.

\textbf{Line and paragraph-level processing} focuses on the structural integrity of the text by removing HTML and JavaScript tags, handling custom structures specific to certain domains, filtering out lines with excessive special characters, and eliminating short or incomplete lines that do not contribute meaningful content.

\textbf{Document-level processing} entails a comprehensive evaluation of each document's relevance and quality. Documents are discarded based on criteria such as insufficient length, predominance of non-Persian content, excessive repetition of words, high proportion of short lines, and the presence of out-of-vocabulary (OOV) words. These filters ensure that only high-quality, relevant, and linguistically coherent documents are retained in the corpus.

After cleaning the documents, we apply a deduplication step to mitigate data redundancy, a crucial aspect of the preprocessing pipeline highlighted in several studies \citep{gao2020pile, penedo2023refinedweb, le2023bloom}.  Utilizing the MinHash algorithm \citep{broder1997minhash}, we efficiently identify and eliminate both exact and near-duplicate documents, thereby enhancing the corpus's uniqueness.

For two manually inspected domains, \href{https://virgool.io/}{Virgool}\footnote{\href{https://virgool.io/}{https://virgool.io/}} and \href{https://en.wikishia.net/}{WikiShia}\footnote{\href{https://en.wikishia.net/}{https://en.wikishia.net/}}, we adopted a tailored processing approach to account for domain-specific characteristics. Virgool's diverse blog posts required relaxed filtering criteria to preserve technical content, while WikiShia's recursive linking and bilingual content deemed for  specialized deduplication and language handling techniques to maintain content integrity and cultural relevance.
\subsection{Crawled Books and Papers}
Data collected from the web alone does not provide sufficient factual or literary content. To enrich our dataset, we also sourced publicly accessible books and academic papers from websites and social media channels. As demonstrated in Figure~\ref{fig:boxplot1}, the box plot of document length distribution clearly shows that books and papers contain significantly longer texts compared to web and social media content, making them more informative and comprehensive. This length, along with the depth of the content, further justifies the inclusion of these sources in our corpus. 

Since most of these sources provide data in PDF format, additional steps were required to convert PDFs into usable text. However, the limited accuracy of Persian OCR systems introduces challenges, particularly when processing PDFs that contain scanned images.

We divided the data from books and papers into two groups, each requiring different processing steps based on the nature of the data: Text-based PDFs and Image-based PDFs (OCR). Just like the data from web, the processing of books and papers involved a combination of document-level, character-level, and line-level operations to ensure data quality, as outlined below.

\subsubsection{Text-based PDFs}
Text-based PDFs primarily include books and academic papers sourced from Telegram channels and Persian websites. The PDFs were converted into text using several Python libraries. To ensure quality, we tested various tools on sample documents and applied low-level heuristic filters to remove corrupted or irrelevant content.

The filtering process involves removing documents with insufficient Persian content, short text lengths, or an excessive use of symbols. This stage ensures that only relevant and high-quality documents are retained. Following this initial filtering, we apply a preprocessing pipeline to address document, character, and line-level inconsistencies, ensuring the text is properly structured. Additional technical details on these steps, including character normalization, watermark removal, and deduplication, are provided in the Appendix~\ref{sec:appendixB}.

\subsubsection{Image-based PDFs (OCR)}
Many papers in our dataset were converted to text using image-based OCR due to the unavailability of text-based PDFs. Given the limitations of Persian OCR, errors were introduced during text extraction. To address this, we filtered out low-quality documents, focusing on those with a high percentage of nonsensical tokens or merged words. As a result, the dataset was refined to include 321,244 documents. The documents were then processed using steps  similar to those applied to web-based crawled data, with additional procedures. Additional information on the OCR-specific filtering methods is provided in the Appendix.

\subsection{Social Media}

Although some books and blogs may include informal Persian text or dialogues, the overall proportion of such data is minimal. The data collected from web-based sources and books generally lacks unstructured or colloquial language. Social media, however, provides a rich source of unstructured and informal linguistic data. To capture this, we gathered Persian-language data from Twitter, as well as public channels and groups from Telegram and Eitaa (an Iranian chat application). After identifying relevant channels and groups, we crawled all associated messages and processed them using the pipeline described for web-based data, with thresholds tailored to social media content. Additional processing steps we applied are outlined below.

Upon examination, we found that shorter messages were mostly replies, often lacking substantive content or containing inappropriate language. These messages were filtered out. We also identified hashtags embedded within the text and at the end of messages. Hashtags within the text were retained to preserve context, while those at the end, frequently related to political or social topics and often irrelevant to the main content, were removed. We employed regular expressions (regex) to remove channel IDs and URLs, ensuring that irrelevant content was minimized.

A notable difference in processing social media data was the deduplication strategy. We observed that many messages from different sources differed only in date or pricing—typically for goods, gold, silver, or cryptocurrencies. To address this, we removed all numeric values and dates before deduplication. After identifying and eliminating duplicate entries, we restored the original content, including numbers and dates, for the final dataset. This method ensured that informative variations were preserved while content containing no new knowledge was removed. 

\subsection{Final Dataset}
\begin{figure*}[t]
  \includegraphics[width=15.5cm]
  {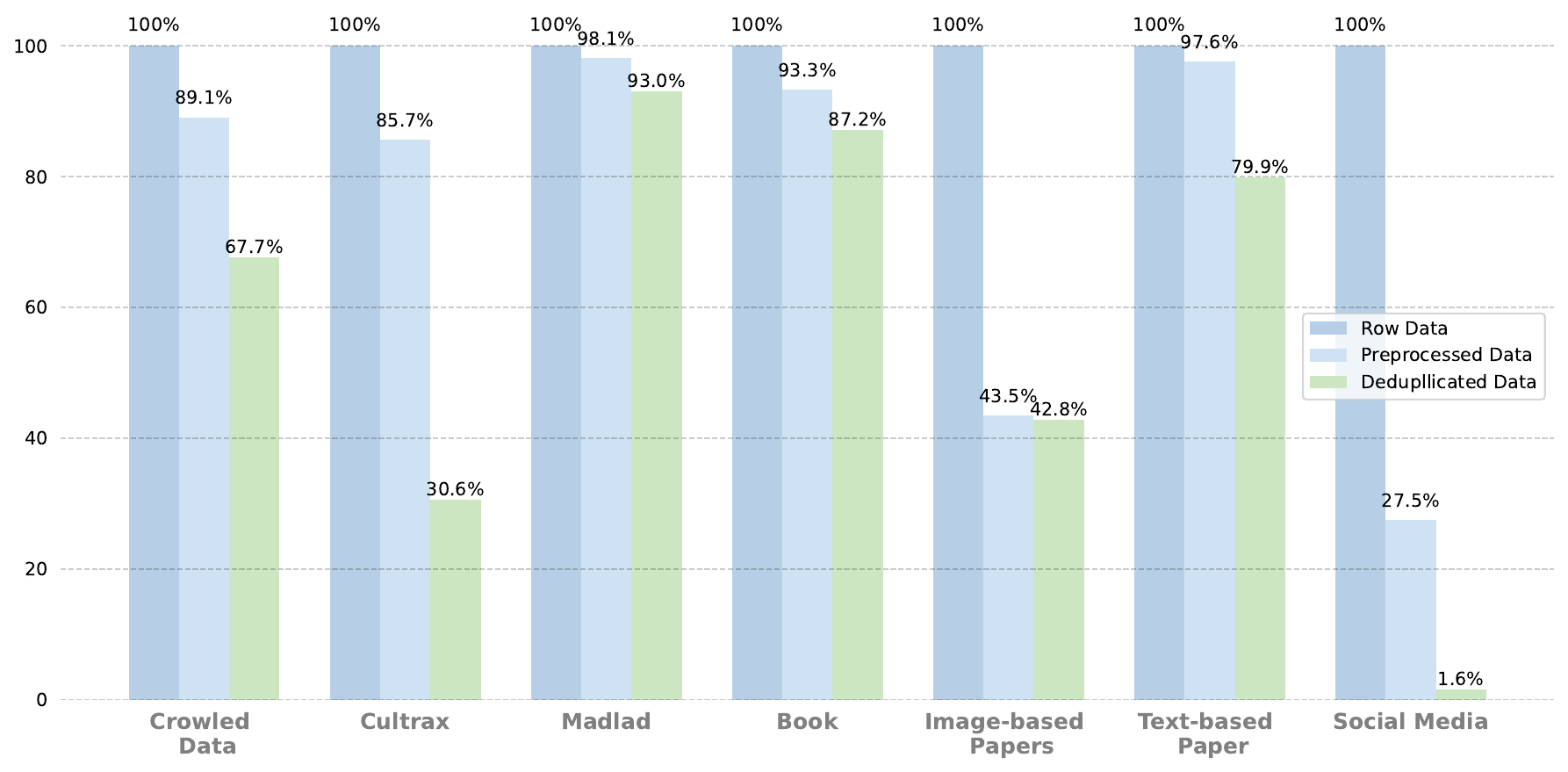}
  \centering
  \caption{Data reduction during preprocessing and deduplication varies significantly across sources. Social media shows the most drastic drop, with just 1.6\% of documents remaining after deduplication, while other sources retain between 56.1\% and 93.3\%. The three bars for each source represent the percentage of documents left after each stage. Overall, about 14\% of the initial documents remain.}
  \label{fig:barplot}
\end{figure*}

Applying the outlined preprocessing steps, including deduplication, resulted in a significant reduction in the number of documents. As illustrated in Figure~\ref{fig:barplot}, the overall document count decreased by an average of \textbf{24\%} after preprocessing, with a further reduction of \textbf{18.83\%} following deduplication.

The largest reduction occurred in social media content, particularly from Twitter and Telegram. Many Twitter posts were short and lacked meaningful content, while Telegram messages were often redundant, brief, and became even less informative after hashtags and links were removed. The special deduplication method we applied also identified many of these messages as duplicates. Although only 1.6\% of social media documents remained after processing, these retained documents were significantly longer, accounting for around 10\% of the total token count from the initial data.

Image-based academic papers also experienced a considerable loss during processing. In this category, the number of documents was nearly halved, as we applied multiple criteria to remove poor-quality documents. In contrast, text-based papers saw minimal loss, with only 2\% of documents eliminated during preprocessing. However, papers in this category contained more duplicates, which contributed to the reduction.

Books had the lowest proportion of document elimination during both preprocessing and deduplication. This reflects the higher quality of book content and the effectiveness of the methods used to extract data from PDF files.

For the web-crawled data, deduplication had a bigger impact than the initial preprocessing, with more documents being removed in this step. Even though we carefully tried to avoid duplicates during crawling, the nature of web crawling—often involving nested links—led to the inclusion of duplicates. Additionally, many news websites repost the same content across different agencies, which shows just how important thorough deduplication is for web-sourced data.

An interesting observation from the bar plot is that, although CulturaX FA \citep{nguyen2023culturax} and Madlad-400 FA \citep{kudugunta2024madlad} claim to have already undergone processing and deduplication, our language-specific preprocessing steps and content-specific deduplication further reduced their size. In Madlad-400 FA, only 7\% of documents were discarded, whereas nearly 70\% of CulturaX FA documents did not meet the qualifications for proper Persian data. This emphasizes the importance of language-specific processing and careful evaluation by native speakers to ensure data quality.

\section{Assessing the Impact of the Matina Corpus}

A large-scale Persian corpus has numerous applications in NLP, including training transformer-based models for tasks such as summarization, sentiment analysis, emotion detection, question answering, sentence embeddings, and text retrieval. Additionally, such corpora play a crucial role in pretraining large language models (LLMs) and generating instructions for LLM post-training. To assess the effectiveness of the Matina Corpus, we conducted experiments on transformer-based model training and continued pretraining of LLMs. This section provides a detailed discussion of these experiments and their outcomes.

\subsection{Masked Language Model Training and Evaluation}
While LLMs have excelled in various NLP tasks such as sentiment analysis and named entity recognition (NER), there remains a need for lightweight models that can be easily fine-tuned for specific tasks and datasets. These models are typically built on transformer-based architectures, particularly masked language models trained on large-scale datasets. 

To address this need, we conducted continual pretraining of masked language models (MLMs), specifically XLM-RoBERTa Large \citep{xlmro}, on 54.69 billion tokens of our dataset. This extensive corpus facilitates the development of high-quality sentence embeddings, further refined by adapting the model into a Sentence-BERT architecture without Next Sentence Prediction (NSP). These enhancements yield more precise semantic representations, significantly improving Persian NLP tasks. By leveraging a well-curated dataset with rigorous preprocessing, our model effectively captures Persian linguistic nuances. 

To evaluate the effectiveness of Matina corpus in training transformer models, we benchmarked out Roberta-based model against existing models using datasets such as \textbf{Arman Emo}, \textbf{Pars-ABSA}, \textbf{PQUAD}, and \textbf{PEYMA}. As shown in Table \ref{tabelMLM}, our model demonstrates substantial performance gains, achieving \textbf{56.54} on \textbf{Arman Emo}, surpassing TookaBERT and AriaBERT, and \textbf{74.92} on \textbf{Pars-ABSA}, highlighting its robustness in aspect-based sentiment analysis. These results validate the impact of our dataset on enhancing Persian NLP performance, particularly within transformer-based architectures.

The success of our MLM underscores the crucial role of high-quality data in pretraining. By capturing Persian linguistic and cultural nuances, our model not only enhances task-specific performance but also advances the goal of developing inclusive and representative language technologies. This approach ensures that underrepresented languages like Persian receive the attention they deserve, fostering more equitable advancements in NLP.

\begin{table*}[t]
    \centering
    \caption{Results of Masked Language Models Evaluation.}
    \begin{tabular}{p{6.4cm}p{2cm}p{2cm}p{1.6cm}p{1.2cm}}
    \hline
        \textbf{Model} 
        & \rotatebox{0}{\textbf{Arman Emo}} 
        & \rotatebox{0}{\textbf{Pars-ABSA}} 
        & \rotatebox{0}{\textbf{PQUAD}} 
        & \rotatebox{0}{\textbf{PEYMA}} \\ \hline
        \textbf{XLM-RoBERTa (ours)} & \textbf{56.54}  & \textbf{74.92}  & 86.82 & \textbf{85.65} \\ 
        \textbf{TookaBERT} \citep{sadraeijavaheri2024tookabert} & 52.87 & 74.65  & 86.73 & 86.09 \\ 
        \textbf{AriaBERT} \citep{ghafouri2023ariabert} & 38.23 & 74.59  & 83.14 & 35.78 \\ 
        \textbf{XLM-RoBERTa} \citep{xlmro} & 32.48 & 74.18 & \textbf{87.6}  & 87.94 \\ 
        \textbf{mBERT} & 6.74 & 68.15 & 85.94 & 65.32 \\ \hline
    \end{tabular}
    \label{tabelMLM}
\end{table*}

\subsection{Large Language Model Pretraining and Evaluation}
\begin{table}[t]
  \centering
  \begin{tabular}{ll}
    \hline
    \textbf{Dataset}           & \textbf{Number of tokens} \\
    \hline
    Social and Politics           &    1.1 B   \\
    Cooking          &    15 M   \\
    \hline
  \end{tabular}
  \caption{\label{table:pretrainData}
    Number of tokens used for LLM continual pretraining. Tokens are counted by the Llama 3.1 \citep{dubey2024llama3.1} tokenizer. 
  }
\end{table}

\begin{figure}[b]
  \includegraphics[trim=0.4cm 0.9cm 0.8cm 0.3cm, width=\columnwidth]{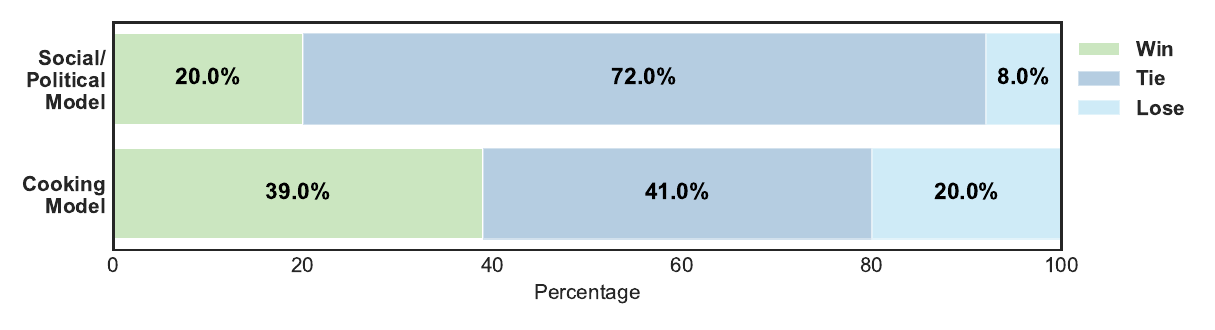}
  \caption{Win rate of pretrained models over models without pretraining.}
  \label{fig:winLose}
\end{figure}

Pretraining is essential for transferring knowledge to LLMs, shaping their linguistic and factual understanding. However, multilingual LLMs often struggle with underrepresented languages like Persian and exhibit cultural biases favoring Western perspectives \cite{cao2023assessing, alkhamissi2024investigating} due to the dominance of English in their training data. This leads to diminished performance in other languages and cultures. Incorporating language-specific data during pretraining can help address this issue.

To evaluate the impact of our dataset on LLM training, we conducted the following experiment. We first tagged our dataset in an unsupervised manner using a procedure similar to InsTag \cite{lu2023instag}, categorizing it into multiple domains. From these, we selected two—social and politics and cooking—and extracted a subset of data from each domain. These domain-specific subsets were then used to train models. The token count for each domain is presented in Table \ref{table:pretrainData}. We then constructed large instruction datasets for these domains and fine-tuned LLaMA 3.2-Instruct 8B using two different approaches: (1) continued pretraining on the domain-specific data followed by instruction tuning, and (2) direct instruction tuning without additional pretraining. To evaluate model performance, we conducted a human evaluation, where annotators ranked model outputs in a win-lose format, indicating which model provided better responses to a held-out evaluation set derived from the instruction dataset.

The evaluation results, shown in Figure \ref{fig:winLose}, indicate that models benefit significantly from pretraining on even a relatively small dataset before instruction tuning. This effect is particularly noticeable in the cooking domain, where the pretrained model was preferred nearly twice as often as the model without pretraining. These findings highlight the effectiveness of the Matina Corpus in improving language models by providing high-quality, domain-specific data. Pretraining on a small, well-curated dataset not only enriches the model’s knowledge but also enhances its alignment with the target language and cultural context.

\section{Conclusion}

In conclusion, the Matina corpus provides a crucial resource for advancing Persian NLP by addressing the limitations of existing datasets in terms of scale and diversity. With 72.9 billion tokens, it enables the training of more advanced and accurate models for tasks such as machine translation, summarization, and large-scale language modeling. We further demonstrate its effectiveness by training and evaluating transformer-based models on key NLP tasks as well as LLM pretraining, highlighting the benefits of high-quality Persian data. By making both the dataset and preprocessing tools publicly available, we aim to support further research and foster collaboration in the development of open-source tools and models for Persian.
\section{Limitations}
While our Persian corpus represents a significant step forward in providing high-quality data, there are several limitations to be noted:

\textbf{Sub-Document Level Redundancies:} Although we applied deduplication at the document level, we did not perform deduplication within documents, meaning there may be redundancies at the sentence or paragraph level. This limitation arises from the high memory and computational resources required to encode and compare sections of all documents. Unfortunately, we did not have the resources necessary to conduct this process at a finer granularity.

\textbf{Sensitive Content and Language:} Despite selecting Persian websites with minimal adult content and removing sensitive data from public datasets, some sensitive material and inappropriate language remain, particularly in social media data. We did not filter out offensive or explicit language, as it reflects real-world language use. However, researchers utilizing the dataset should be mindful of this content when applying it in their work.

\textbf{Residual Irrelevant Data:} While we inspected samples from all data sources and employed various heuristics and filtering functions to remove irrelevant content, such as links, hashtags, advertisements, and tags, some may have evaded our processes. These elements are generally considered noise given the large scale of the dataset but may need to be addressed for more specialized use cases.

These limitations highlight potential areas for improvement, especially for projects with specific needs regarding data quality and sensitivity.
\section*{Acknowledgements}
We would like to sincerely thank Mohammad Ebrahimnezhadian for his invaluable assistance in converting PDFs to usable text formats. Additionally, we would like to extend our appreciation to the National Artificial Intelligence Organization for their generous financial support, which was instrumental in the completion of this work.

\bibliography{acl_latex}

\begin{thebibliography}{42}
\providecommand{\natexlab}[1]{#1}

\bibitem[{AlKhamissi et~al.(2024)AlKhamissi, ElNokrashy, AlKhamissi, and Diab}]{alkhamissi2024investigating}
Badr AlKhamissi, Muhammad ElNokrashy, Mai AlKhamissi, and Mona Diab. 2024.
\newblock Investigating cultural alignment of large language models.
\newblock \emph{arXiv preprint arXiv:2402.13231}.

\bibitem[{Athiwaratkun et~al.(2018)Athiwaratkun, Wilson, and Anandkumar}]{athiwaratkun2018fasttext}
Ben Athiwaratkun, Andrew~Gordon Wilson, and Anima Anandkumar. 2018.
\newblock Probabilistic fasttext for multi-sense word embeddings.
\newblock \emph{arXiv preprint arXiv:1806.02901}.

\bibitem[{Bai et~al.(2023)Bai, Bai, Chu, Cui, Dang, Deng, Fan, Ge, Han, Huang et~al.}]{bai2023qwen}
Jinze Bai, Shuai Bai, Yunfei Chu, Zeyu Cui, Kai Dang, Xiaodong Deng, Yang Fan, Wenbin Ge, Yu~Han, Fei Huang, et~al. 2023.
\newblock Qwen technical report.
\newblock \emph{arXiv preprint arXiv:2309.16609}.

\bibitem[{Broder(1997)}]{broder1997minhash}
Andrei~Z Broder. 1997.
\newblock On the resemblance and containment of documents.
\newblock In \emph{Proceedings. Compression and Complexity of SEQUENCES 1997 (Cat. No. 97TB100171)}, pages 21--29. IEEE.

\bibitem[{Brown(2020)}]{brown2020language}
Tom~B Brown. 2020.
\newblock Language models are few-shot learners.
\newblock \emph{arXiv preprint arXiv:2005.14165}.

\bibitem[{Cao et~al.(2023)Cao, Zhou, Lee, Cabello, Chen, and Hershcovich}]{cao2023assessing}
Y~Cao, L~Zhou, S~Lee, L~Cabello, M~Chen, and D~Hershcovich. 2023.
\newblock Assessing cross-cultural alignment between chatgpt and human societies: An empirical study. arxiv.
\newblock \emph{Preprint posted online on March}, 31.

\bibitem[{Computer(2023)}]{together2023redpajama}
Together Computer. 2023.
\newblock \href {https://github.com/togethercomputer/RedPajama-Data} {Redpajama: an open dataset for training large language models}.

\bibitem[{Conneau(2019{\natexlab{a}})}]{conneau2019unsupervised}
A~Conneau. 2019{\natexlab{a}}.
\newblock Unsupervised cross-lingual representation learning at scale.
\newblock \emph{arXiv preprint arXiv:1911.02116}.

\bibitem[{Conneau(2019{\natexlab{b}})}]{conneau2019cc100}
A~Conneau. 2019{\natexlab{b}}.
\newblock Unsupervised cross-lingual representation learning at scale.
\newblock \emph{arXiv preprint arXiv:1911.02116}.

\bibitem[{Conneau et~al.(2020)Conneau, Khandelwal, Goyal, Chaudhary, Wenzek, Guzmán, Grave, Ott, Zettlemoyer, and Stoyanov}]{xlmro}
Alexis Conneau, Kartikay Khandelwal, Naman Goyal, Vishrav Chaudhary, Guillaume Wenzek, Francisco Guzmán, Edouard Grave, Myle Ott, Luke Zettlemoyer, and Veselin Stoyanov. 2020.
\newblock \href {https://arxiv.org/abs/1911.02116} {Unsupervised cross-lingual representation learning at scale}.
\newblock \emph{Preprint}, arXiv:1911.02116.

\bibitem[{Crawl(2008)}]{commoncrawl}
Common Crawl. 2008.
\newblock \href {https://commoncrawl.org} {Common crawl corpus}.
\newblock Accessed: 2024-09-28.

\bibitem[{Devlin(2018)}]{devlin2018bert}
Jacob Devlin. 2018.
\newblock Bert: Pre-training of deep bidirectional transformers for language understanding.
\newblock \emph{arXiv preprint arXiv:1810.04805}.

\bibitem[{Dubey et~al.(2024)Dubey, Jauhri, Pandey, Kadian, Al-Dahle, Letman, Mathur, Schelten, Yang, Fan et~al.}]{dubey2024llama3.1}
Abhimanyu Dubey, Abhinav Jauhri, Abhinav Pandey, Abhishek Kadian, Ahmad Al-Dahle, Aiesha Letman, Akhil Mathur, Alan Schelten, Amy Yang, Angela Fan, et~al. 2024.
\newblock The llama 3 herd of models.
\newblock \emph{arXiv preprint arXiv:2407.21783}.

\bibitem[{Gao et~al.(2020)Gao, Biderman, Black, Golding, Hoppe, Foster, Phang, He, Thite, Nabeshima et~al.}]{gao2020pile}
Leo Gao, Stella Biderman, Sid Black, Laurence Golding, Travis Hoppe, Charles Foster, Jason Phang, Horace He, Anish Thite, Noa Nabeshima, et~al. 2020.
\newblock The pile: An 800gb dataset of diverse text for language modeling.
\newblock \emph{arXiv preprint arXiv:2101.00027}.

\bibitem[{Ghafouri et~al.(2023)Ghafouri, Abbasi, and Naderi}]{ghafouri2023ariabert}
Arash Ghafouri, Mohammad~Amin Abbasi, and Hassan Naderi. 2023.
\newblock Ariabert: A pre-trained persian bert model for natural language understanding.

\bibitem[{Glockner et~al.(2018)Glockner, Shwartz, and Goldberg}]{glockner2018snli}
Max Glockner, Vered Shwartz, and Yoav Goldberg. 2018.
\newblock \href {https://arxiv.org/abs/1805.02266} {Breaking nli systems with sentences that require simple lexical inferences}.
\newblock \emph{Preprint}, arXiv:1805.02266.

\bibitem[{Kashefi(2020)}]{kashefi2020mizanlargepersianenglishparallel}
Omid Kashefi. 2020.
\newblock \href {https://arxiv.org/abs/1801.02107} {Mizan: A large persian-english parallel corpus}.
\newblock \emph{Preprint}, arXiv:1801.02107.

\bibitem[{Khansari and Shamsfard(2021)}]{khansari2021hmblogsbiggeneralpersian}
Hamzeh~Motahari Khansari and Mehrnoush Shamsfard. 2021.
\newblock \href {https://arxiv.org/abs/2111.02362} {Hmblogs: A big general persian corpus}.
\newblock \emph{Preprint}, arXiv:2111.02362.

\bibitem[{Kudugunta et~al.(2024)Kudugunta, Caswell, Zhang, Garcia, Xin, Kusupati, Stella, Bapna, and Firat}]{kudugunta2024madlad}
Sneha Kudugunta, Isaac Caswell, Biao Zhang, Xavier Garcia, Derrick Xin, Aditya Kusupati, Romi Stella, Ankur Bapna, and Orhan Firat. 2024.
\newblock Madlad-400: A multilingual and document-level large audited dataset.
\newblock \emph{Advances in Neural Information Processing Systems}, 36.

\bibitem[{Lauren{\c{c}}on et~al.(2022)Lauren{\c{c}}on, Saulnier, Wang, Akiki, Villanova~del Moral, Le~Scao, Von~Werra, Mou, Gonz{\'a}lez~Ponferrada, Nguyen et~al.}]{laurenccon2022roots}
Hugo Lauren{\c{c}}on, Lucile Saulnier, Thomas Wang, Christopher Akiki, Albert Villanova~del Moral, Teven Le~Scao, Leandro Von~Werra, Chenghao Mou, Eduardo Gonz{\'a}lez~Ponferrada, Huu Nguyen, et~al. 2022.
\newblock The bigscience roots corpus: A 1.6 tb composite multilingual dataset.
\newblock \emph{Advances in Neural Information Processing Systems}, 35:31809--31826.

\bibitem[{Le~Scao et~al.(2023)Le~Scao, Fan, Akiki, Pavlick, Ili{\'c}, Hesslow, Castagn{\'e}, Luccioni, Yvon, Gall{\'e} et~al.}]{le2023bloom}
Teven Le~Scao, Angela Fan, Christopher Akiki, Ellie Pavlick, Suzana Ili{\'c}, Daniel Hesslow, Roman Castagn{\'e}, Alexandra~Sasha Luccioni, Fran{\c{c}}ois Yvon, Matthias Gall{\'e}, et~al. 2023.
\newblock Bloom: A 176b-parameter open-access multilingual language model.

\bibitem[{Leskovec et~al.(2020)Leskovec, Rajaraman, and Ullman}]{leskovec2020lsh}
Jure Leskovec, Anand Rajaraman, and Jeffrey~David Ullman. 2020.
\newblock \emph{Mining of massive data sets}.
\newblock Cambridge university press.

\bibitem[{Lu et~al.(2023)Lu, Yuan, Yuan, Lin, Lin, Tan, Zhou, and Zhou}]{lu2023instag}
Keming Lu, Hongyi Yuan, Zheng Yuan, Runji Lin, Junyang Lin, Chuanqi Tan, Chang Zhou, and Jingren Zhou. 2023.
\newblock \# instag: Instruction tagging for analyzing supervised fine-tuning of large language models.
\newblock In \emph{The Twelfth International Conference on Learning Representations}.

\bibitem[{Muennighoff et~al.(2022)Muennighoff, Tazi, Magne, and Reimers}]{muennighoff2022massiveText}
Niklas Muennighoff, Nouamane Tazi, Lo{\"\i}c Magne, and Nils Reimers. 2022.
\newblock Mteb: Massive text embedding benchmark.
\newblock \emph{arXiv preprint arXiv:2210.07316}.

\bibitem[{Narayan et~al.(2018)Narayan, Cohen, and Lapata}]{narayan2018xsum}
Shashi Narayan, Shay~B. Cohen, and Mirella Lapata. 2018.
\newblock \href {https://arxiv.org/abs/1808.08745} {Don't give me the details, just the summary! topic-aware convolutional neural networks for extreme summarization}.
\newblock \emph{Preprint}, arXiv:1808.08745.

\bibitem[{Nguyen et~al.(2023)Nguyen, Van~Nguyen, Lai, Man, Ngo, Dernoncourt, Rossi, and Nguyen}]{nguyen2023culturax}
Thuat Nguyen, Chien Van~Nguyen, Viet~Dac Lai, Hieu Man, Nghia~Trung Ngo, Franck Dernoncourt, Ryan~A Rossi, and Thien~Huu Nguyen. 2023.
\newblock Culturax: A cleaned, enormous, and multilingual dataset for large language models in 167 languages.
\newblock \emph{arXiv preprint arXiv:2309.09400}.

\bibitem[{Penedo et~al.(2024)Penedo, Kydl{\'\i}{\v{c}}ek, Lozhkov, Mitchell, Raffel, Von~Werra, Wolf et~al.}]{penedo2024fineweb}
Guilherme Penedo, Hynek Kydl{\'\i}{\v{c}}ek, Anton Lozhkov, Margaret Mitchell, Colin Raffel, Leandro Von~Werra, Thomas Wolf, et~al. 2024.
\newblock The fineweb datasets: Decanting the web for the finest text data at scale.
\newblock \emph{arXiv preprint arXiv:2406.17557}.

\bibitem[{Penedo et~al.(2023)Penedo, Malartic, Hesslow, Cojocaru, Cappelli, Alobeidli, Pannier, Almazrouei, and Launay}]{penedo2023refinedweb}
Guilherme Penedo, Quentin Malartic, Daniel Hesslow, Ruxandra Cojocaru, Alessandro Cappelli, Hamza Alobeidli, Baptiste Pannier, Ebtesam Almazrouei, and Julien Launay. 2023.
\newblock The refinedweb dataset for falcon llm: outperforming curated corpora with web data, and web data only.
\newblock \emph{arXiv preprint arXiv:2306.01116}.

\bibitem[{Radford(2018)}]{radford2018improving}
Alec Radford. 2018.
\newblock Improving language understanding by generative pre-training.

\bibitem[{Raffel et~al.(2020)Raffel, Shazeer, Roberts, Lee, Narang, Matena, Zhou, Li, and Liu}]{raffel2020c4}
Colin Raffel, Noam Shazeer, Adam Roberts, Katherine Lee, Sharan Narang, Michael Matena, Yanqi Zhou, Wei Li, and Peter~J Liu. 2020.
\newblock Exploring the limits of transfer learning with a unified text-to-text transformer.
\newblock \emph{Journal of machine learning research}, 21(140):1--67.

\bibitem[{Sabeti et~al.(2018)Sabeti, Firouzjaee, Choobbasti, Najafabadi, and Vaheb}]{sabeti2018mirastext}
Behnam Sabeti, Hossein~Abedi Firouzjaee, Ali~Janalizadeh Choobbasti, SHE~Mortazavi Najafabadi, and Amir Vaheb. 2018.
\newblock Mirastext: An automatically generated text corpus for persian.
\newblock In \emph{Proceedings of the eleventh international conference on language resources and evaluation (LREC 2018)}.

\bibitem[{Sabouri et~al.(2022)Sabouri, Rahmati, Gooran, and Sameti}]{naab}
Sadra Sabouri, Elnaz Rahmati, Soroush Gooran, and Hossein Sameti. 2022.
\newblock \href {https://arxiv.org/abs/2208.13486} {naab: A ready-to-use plug-and-play corpus for farsi}.
\newblock \emph{Preprint}, arXiv:2208.13486.

\bibitem[{SadraeiJavaheri et~al.(2024)SadraeiJavaheri, Moghaddaszadeh, Molazadeh, Naeiji, Aghababaloo, Rafiee, Amirmahani, Abedini, Sheikhi, and Salehoof}]{sadraeijavaheri2024tookabert}
MohammadAli SadraeiJavaheri, Ali Moghaddaszadeh, Milad Molazadeh, Fariba Naeiji, Farnaz Aghababaloo, Hamideh Rafiee, Zahra Amirmahani, Tohid Abedini, Fatemeh~Zahra Sheikhi, and Amirmohammad Salehoof. 2024.
\newblock Tookabert: A step forward for persian nlu.
\newblock \emph{arXiv preprint arXiv:2407.16382}.

\bibitem[{Su{\'a}rez et~al.(2019)Su{\'a}rez, Sagot, and Romary}]{oscar}
Pedro Javier~Ortiz Su{\'a}rez, Beno{\^\i}t Sagot, and Laurent Romary. 2019.
\newblock Asynchronous pipeline for processing huge corpora on medium to low resource infrastructures.
\newblock In \emph{7th Workshop on the Challenges in the Management of Large Corpora (CMLC-7)}. Leibniz-Institut f{\"u}r Deutsche Sprache.

\bibitem[{Targoman(2022)}]{targoman}
Targoman. 2022.
\newblock \href {https://oss.targoman.ir/} {Targoman dataset}.

\bibitem[{Tiedemann(2012)}]{tiedemann-2012-parallel}
J{\"o}rg Tiedemann. 2012.
\newblock \href {http://www.lrec-conf.org/proceedings/lrec2012/pdf/463_Paper.pdf} {Parallel data, tools and interfaces in {OPUS}}.
\newblock In \emph{Proceedings of the Eighth International Conference on Language Resources and Evaluation ({LREC}'12)}, pages 2214--2218, Istanbul, Turkey. European Language Resources Association (ELRA).

\bibitem[{Touvron et~al.(2023)Touvron, Lavril, Izacard, Martinet, Lachaux, Lacroix, Rozi{\`e}re, Goyal, Hambro, Azhar et~al.}]{touvron2023llama}
Hugo Touvron, Thibaut Lavril, Gautier Izacard, Xavier Martinet, Marie-Anne Lachaux, Timoth{\'e}e Lacroix, Baptiste Rozi{\`e}re, Naman Goyal, Eric Hambro, Faisal Azhar, et~al. 2023.
\newblock Llama: Open and efficient foundation language models.
\newblock \emph{arXiv preprint arXiv:2302.13971}.

\bibitem[{Vaswani(2017)}]{vaswani2017attention}
A~Vaswani. 2017.
\newblock Attention is all you need.
\newblock \emph{Advances in Neural Information Processing Systems}.

\bibitem[{Wang et~al.(2019)Wang, Singh, Michael, Hill, Levy, and Bowman}]{wang2019glue}
Alex Wang, Amanpreet Singh, Julian Michael, Felix Hill, Omer Levy, and Samuel~R. Bowman. 2019.
\newblock \href {https://arxiv.org/abs/1804.07461} {Glue: A multi-task benchmark and analysis platform for natural language understanding}.
\newblock \emph{Preprint}, arXiv:1804.07461.

\bibitem[{Wenzek et~al.(2019)Wenzek, Lachaux, Conneau, Chaudhary, Guzm{\'a}n, Joulin, and Grave}]{wenzek2019ccnet}
Guillaume Wenzek, Marie-Anne Lachaux, Alexis Conneau, Vishrav Chaudhary, Francisco Guzm{\'a}n, Armand Joulin, and Edouard Grave. 2019.
\newblock Ccnet: Extracting high quality monolingual datasets from web crawl data.
\newblock \emph{arXiv preprint arXiv:1911.00359}.

\bibitem[{Xue(2020)}]{xue2020mt5}
L~Xue. 2020.
\newblock mt5: A massively multilingual pre-trained text-to-text transformer.
\newblock \emph{arXiv preprint arXiv:2010.11934}.

\bibitem[{Yang et~al.(2024)Yang, Yang, Hui, Zheng, Yu, Zhou, Li, Li, Liu, Huang et~al.}]{yang2024qwen2}
An~Yang, Baosong Yang, Binyuan Hui, Bo~Zheng, Bowen Yu, Chang Zhou, Chengpeng Li, Chengyuan Li, Dayiheng Liu, Fei Huang, et~al. 2024.
\newblock Qwen2 technical report.
\newblock \emph{arXiv preprint arXiv:2407.10671}.

\end{thebibliography}

\appendix
\renewcommand{\thesection}{\Alph{section}}
\renewcommand{\thesubsection}{\Alph{section}.\arabic{subsection}}

\section*{Appendix}

\section{Details of Web-Based Document Processing Pipeline}
\label{sec:appendixA}
\subsection{Character-level Processing}
Character-level processing is the initial step in our preprocessing pipeline, aimed at standardizing and cleaning the text at the most granular level. This stage involves several key operations:

\begin{enumerate}
    \item Unicode Normalization: We convert all characters to their Persian equivalents, and remove Arabic I'rab marks. We then normalize space and tab characters to the standard keyboard space, with exceptions made for the half-space character used in specific Persian words.
    
    \item Symbol and Number Mapping: We map symbols and numbers not belonging to the English, Arabic, or Persian character sets to their Persian equivalents using the \href{https://github.com/arushadev/piraye}{Piraye} library. This is to ensure language consistency in the dataset.
    
    \item Repeated Characters: We identify any character repeated more than three times in sequence, typically used for emphasis, and truncate it to three occurrences to maintain readability and consistency.
    
    \item Newline Normalization: We merge consecutive newlines, including those with spaces or tabs, to standardize line breaks across documents.
    
    \item Non-standard Unicode Removal: By taking multiple samples from the data we found that there are chracters within the text that are not standard. We then detect and remove these non-standard Unicode characters, such as special emojis or corrupted symbols (e.g., bordered question marks) based on our predefined criteria. 
\end{enumerate}

\subsection{Line and Paragraph-level Processing}
Once character-level normalization is complete, we focus on the structural elements of the text. This stage involves:
\begin{enumerate}
    \item HTML and JavaScript Tag Removal: We identify lines containing HTML or JavaScript tags and functions using regular expressions and replace them with newlines.

    \item Custom Structures Handling: We inspected that some domains include unique tag structures that do not follow the format of standard tags (JavaScript and HTML) which are not captured by regular expressions. We identify and remove these using structures. 

    \item Special Character Ratio Filtering: We calculate the ratio of special characters (e.g., emojis, symbols, numbers) to total characters in each line. Lines exceeding a 0.85 ratio are removed, particularly targeting lines corrupted during text extraction, such as tables or formulas.

    \item Short Line Removal: We inspected that certain sources contain incomplete or irrelevant information in the few short lines at the start of the content. We therefore remove lines shorter these specific sources.
\end{enumerate}

\subsection{Document-level Processing}
The final stage involves document-level processing. We treat documents as a whole and remove those that meet any of the following criteria: (we refer to words as space-separated text sequences that are neither a number nor a symbol)
\begin{itemize}
    \item Short Length Filtering: Documents shorter than 30 words are removed, as they are either corrupted or devoid of useful information.

    \item Non-Persian Content Removal: Documents where over 50\% of characters are non-Persian are eliminated to maintain linguistic consistency and relevance.

    \item Repeated Words Elimination: Documents where more than 50\% of the words are identical are eliminated, targeting pages that use SEO techniques or lack informative content.

    \item Short Lines Proportion Filtering: Documents with over 50\% of lines shorter than 15 words are discarded, as they typically consist of lists or content tables.

    \item Out-of-Vocabulary (OOV) Words Filtering: Specifically for the CulturaX \citep{nguyen2023culturax} dataset, documents containing more than 2.5\% OOV words are removed to exclude irrelevant content such as code fragments or corrupted text.
\end{itemize} 

Finally, we eliminate any repeated empty newlines resulted from the removal of lines or paragraphs to maintain the document's structural integrity.

\subsection{Deduplication Process}
To address data redundancy, we leverage the MinHash algorithm \citep{broder1997minhash}, a well-established technique for efficient similarity detection in large collections of text. The deduplication pipeline consists of the following steps:\citep{broder1997minhash}. The process involves several steps:
\begin{enumerate}
    \item Text Normalization: We normalize Text within all documents by unifying recurring elements like days of the week and removing numbers and symbols. This normalization step is particularly crucial for content from websites that repost similar material daily. By handling these elements, we aim to reduce semantic duplicates.

    \item Tokenization and Hashing: We tokenize each document into 13-grams, and hash values are computed using 128 distinct hashing functions to capture text patterns.

    \item LeanMeanHash Compression: We then segment the hash values into eight sliding windows and processed using the LeanMeanHash algorithm, which compresses the hash signatures for efficient storage and comparison.

    \item Graph-based Similarity Detection: Finally, we construct a graph in which each node represents a document, and edges connect nodes based on hash similarity. By identifying connected components within this graph, only one representative document per component is retained, effectively removing duplicates and near-duplicates.
\end{enumerate}

This deduplication strategy ensures a significant reduction in redundant data, enhancing the corpus's quality and uniqueness, and facilitating better model generalization by preventing overfitting on repeated content.

\subsection{Domain-specific Processing}
Since \href{https://virgool.io/}{Virgool} and \href{https://en.wikishia.net/}{WikiShia} domains contain highly relevant content related to Persian culture and religion, it is  necessary to modify our standard preprocessing pipeline to avoid information loss. We perform the following specialized preprocessings.

For Virgool, which primarily features blog posts on diverse topics, including programming languages and mathematical content, applying the standard preprocessing thresholds resulted in the removal of valuable content. To address this, we relaxed certain filtering criteria: 
\begin{itemize}
    \item By pass the removal of numbers and symbols to preserve technical content.
    \item Incorporate more complex regular expressions to accurately detect and remove residual HTML tags or functions that were not filtered out by the standard pipeline.
    \item Adjust the ratio of Persian stopwords to lower values, and the threshold for the proportion of short lines (in relation to the total number of lines) was increased, ensuring the retention of concise but informative posts.
    \item Employed a privacy-preserving step to remove any personal data found in public blogs, even though the blogs are publicly accessible. This aspect of our pipeline will be discussed in detail in the subsequent section.
\end{itemize}

Another unique challenge with WikiShia was the significant presence of Arabic text, particularly due to references to the Quran and Arabic scholarly sources. To address this, we adjusted our processing thresholds: we increased the tolerance for Arabic stopwords while simultaneously lowering the threshold for Persian stopwords. This adjustment allowed us to better capture the bilingual nature of the content.

For WikiShia which includes bilingual content and presents challenges related to content duplication, we performe the following:

\begin{itemize}
    \item Content Duplication: our recursive crawling process exposed a significant issue of content duplication. Multiple URLs often corresponded to the same page, differing only by a minor subheading. Additionally, the site includes detailed descriptions of events associated with specific dates, resulting in multiple unique URLs hosting nearly identical content tied to calendar events.
    To address this, we employed an exact-match deduplication strategy using MinHashLSH \citep{leskovec2020lsh}. Unlike our standard deduplication pipeline, we opted not to normalize or remove dates, numbers, or references to specific days of the week, as these elements are critical for preserving the chronological and cultural relevance of the content. By applying this approach, we were able to eliminate documents with a similarity threshold of 98\% or higher.

    \item Bilingual Content Handling: Another unique challenge with WikiShia was the significant presence of Arabic text, particularly due to references to the Quran and Arabic scholarly sources. To address this, we adjusted our processing thresholds. The tolerance for Arabic stopwords was increased, while the threshold for Persian stopwords was lowered, effectively capturing the bilingual nature of the content.
\end{itemize}

\begin{figure}[t]
  \includegraphics[width=\columnwidth]{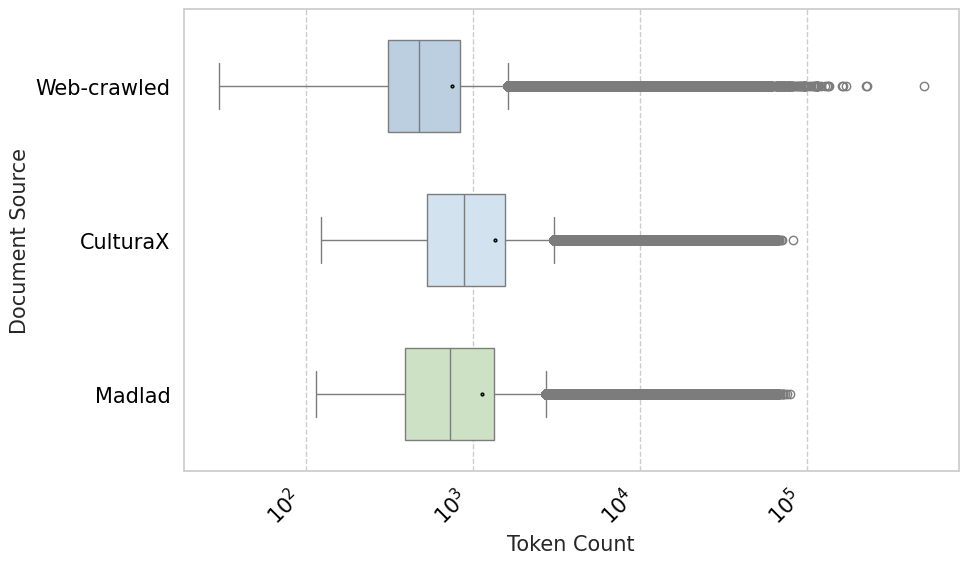}
  \caption{Document Length Distribution For Web-based Crawled Data}
  \label{fig:baxplot3}
\end{figure}

The boxplot in Figure~\ref{fig:baxplot3} illustrates the token count distribution across three document sources we had for web-based data. The results are in tokens by the Llama3.1 \citep{dubey2024llama3.1} and after the application of our comprehensive preprocessing pipeline and deduplication. Data crawled by the team, named Web-crawled, show the widest range, with a median around 1000 tokens and some documents extending beyond \(10^5\) tokens. Madlad exhibits a slightly narrower distribution but still maintains substantial variation. CulturaX demonstrates the most compact distribution, with a lower median and maximum token count. These distributions highlight the success of our preprocessing in maintaining diversity while standardizing document lengths. The presence of outliers, particularly in the Web-crawled and Madlad sources, indicates that our pipeline preserves some longer, potentially information-rich documents. This final data composition ensures a balance between consistency and variety, crucial for robust model training and generalization.

\section{Details of Book and Paper Processing Pipeline}
For data extraction and OCR conversion, we utilized a range of Python libraries, including Selenium\footnote{\url{https://selenium-python.readthedocs.io/}}, BeautifulSoup\footnote{\url{https://beautiful-soup-4.readthedocs.io/en/latest/}}, and Pytesseract\footnote{\url{https://github.com/madmaze/pytesseract}}. Text-based PDFs were converted using lightweight tools such as pdf2image\footnote{\url{https://github.com/Belval/pdf2image}}, while image-based PDFs required more advanced processing with Pytesseract and Fitz\footnote{\url{https://github.com/pymupdf/PyMuPDF}}. To improve accuracy, we employed an iterative approach, applying multiple tools to the same documents and manually inspecting those with errors before refining the extraction process.

\label{sec:appendixB}
\subsection{Text-based PDFs: Detailed Processing}
After removing corrupted or non-Persian documents, we apply a 3-stage processing pipeline involving document-level, character-level and line-level processing. Unlike documents from web, we first apply the document-level processing to avoid redundant processing. 

\subsubsection{Document-level Processing}
In the first stage, we applied document-level processing, where a document was viewed holistically. If it met any of the following criteria, it was eliminated: 
\begin{itemize} 
    \item Documents with fewer than 150 space-separated words. 
    \item Documents containing less than 50\% Persian characters. 
    \item Documents with an average word length of fewer than 3 characters or greater than 10 characters. 
    \item Documents with a numeric or symbolic character ratio exceeding 0.8. 
    \item Documents where over 80\% of the lines were considered short, defined as containing fewer than four space-separated words. 
    \item Documents where fewer than 10\% of the words were Persian or Arabic stopwords. 
\end{itemize}

\subsubsection{Character-level Processing}
Given that many of the books contained long Arabic text, which needed to be preserved, we only normalized non-Arabic, non-English, and non-Persian characters and symbols to their Persian format. We did not remove I'rab (diacritics). Standard procedures, such as replacing consecutive repeated characters, normalizing newlines, and removing non-standard Unicode characters, were applied as in previous section, though with additional Unicode characters added to the filtering set. Furthermore, nonsensical patterns detected in the text, which added no value and increased noise, were removed. These patterns included: 
\begin{itemize} 
    \item Website links to the source of the document. 
    \item Repeated occurrences of the book's title at the top or bottom of pages. 
    \item Page numbers in various forms, such as \FR{صفحه۱}, \FR{صفحه ۱}, \FR{صفحه۱ از ۲۰۰}, \FR{ص۱}, \FR{صفحه ۱ از کتاب ...}, etc. 
    \item Tags related to cover pages. \item Errors or tags related to multimedia, such as \texttt{'Your browser does not support the audio tag.'} 
    \item Images or tables converted to 'UNKNOWN' strings. 
    \item Personal information, such as phone numbers, email addresses, account numbers, and credit card numbers (e.g., Shaba numbers), which were found at the end of some books and at the beginning of papers. 
\end{itemize}

\subsubsection{Line-level Processing}
Following character-level processing, we performed line-level processing to remove lines that contained formulas or tables that were corrupted during the conversion from PDF to text. As part of this stage, the following types of lines or paragraphs were removed: 
\begin{itemize} 
    \item Lines with a numeric character ratio exceeding 0.8. 
    \item Lines with a symbolic character ratio exceeding 0.8. 
    \item Lines that were repeated multiple times within the document, which often included hidden watermarks or the repeated mention of the book's title. 
\end{itemize}

\subsubsection{Deduplication}
To avoid redundancy, a deduplication process was applied using MinHash and Locality-Sensitive Hashing (LSH). We deduplicated documents within each source, ensuring that only unique documents were retained.

\subsection{Image-based PDFs (OCR): Detailed Processing}
For OCR-processed documents, the primary issue was the introduction of errors during text extraction. To mitigate this, we employed the following steps:
\begin{enumerate}
    \item Removed content preceding the keywords section, which was often corrupted, using regex patterns to detect specific document structures.
    \item Removed documents with more than 5\% out-of-vocabulary tokens.
    \item REmoved papers containing more than 10 words exceeding 15 characters, indicative of merged words.
\end{enumerate}
Although some OCR-generated text still contains minor issues, such as occasional word merging, these are manageable with model tokenizers and do not significantly affect overall context and understanding.

\begin{figure}[th]
  \includegraphics[width=\columnwidth]{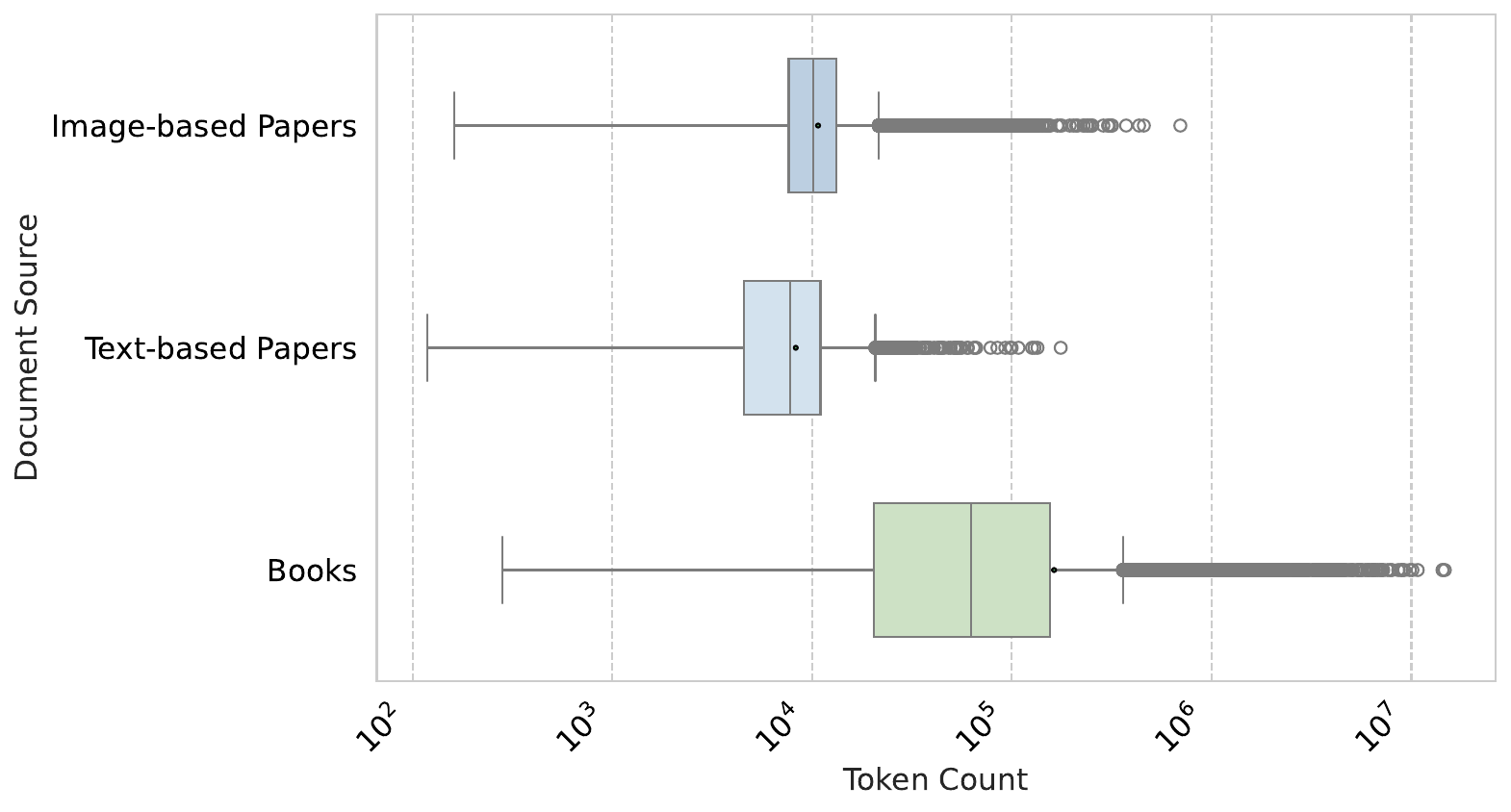}
  \caption{Document Length Distribution For Crawled Books and Papers}
  \label{fig:baxplot2}
\end{figure}

The boxplot in Figure~\ref{fig:baxplot2} shows the token count distribution across different document sources. Books have a notably higher median token count and broader range compared to papers. Both image-based and text-based papers display lower token counts with numerous outliers, indicating diverse token lengths. Text-based papers have a lower median as they contain paper summaries as well as internal papers. Image-based papers also contain high-quality and longer scientific documents. 

\end{document}